\documentclass[conference]{IEEEtran}
\IEEEoverridecommandlockouts

\usepackage{color}
\usepackage{pgfplots}
\pgfplotsset{width=7cm,compat=1.8}

\usepackage{cite}
\usepackage{amsmath, amssymb, amsfonts}
\usepackage{algorithmic}
\usepackage{graphicx}
\usepackage{textcomp}
\usepackage{xcolor}
\usepackage{comment}
\usepackage{subfig}
\usepackage{hyperref}

\newcommand\ddfrac[2]{\frac{\displaystyle #1}{\displaystyle #2}}

\begin{document}

\title{CamLoc: Pedestrian Location Detection from Pose Estimation on Resource-constrained Smart-cameras}

\author{

\IEEEauthorblockN{Adrian Cosma} 
\IEEEauthorblockA{\textit{University Politehnica of Bucharest} \\  
\textit{ioan\_adrian.cosma@stud.acs.upb.ro}}  
\and

\IEEEauthorblockN{Ion Emilian Radoi} 
\IEEEauthorblockA{\textit{University Politehnica of Bucharest} \\  
\textit{emilian.radoi@cs.pub.ro}}  
\and

\IEEEauthorblockN{Valentin Radu} 
\IEEEauthorblockA{\textit{The University of Edinburgh} \\  
\textit{valentin.radu@ed.ac.uk}}  
}
\maketitle

\begin{abstract}

Recent advancements in energy-efficient hardware technology is driving the exponential growth we are experiencing in the Internet of Things (IoT) space, with more pervasive computations being performed near to data generation sources. A range of intelligent devices and applications performing local detection is emerging (activity recognition, fitness monitoring, etc.) bringing with them obvious advantages such as reducing detection latency for improved interaction with devices and safeguarding user data by not leaving the device. Video processing holds utility for many emerging applications and data labelling in the IoT space. However, performing this video processing with deep neural networks at the edge of the Internet is not trivial. In this paper we show that pedestrian location estimation using deep neural networks is achievable on fixed cameras with limited compute resources. Our approach uses pose estimation from key body points detection to extend pedestrian skeleton when whole body not in image (occluded by obstacles or partially outside of frame), which achieves better location estimation performance (infrence time and memory footprint) compared to fitting a bounding box over pedestrian and scaling. We collect a sizable dataset comprising of over 2100 frames in videos from one and two surveillance cameras pointing from different angles at the scene, and annotate each frame with the exact position of person in image, in 42 different scenarios of activity and occlusion. We compare our pose estimation based location detection with a popular detection algorithm, YOLOv2, for overlapping bounding-box generation, our solution achieving faster inference time (15x speedup) at half the memory footprint, within resource capabilities on embedded devices, which demonstrate that CamLoc is an efficient solution for location estimation in videos on smart-cameras.

\end{abstract}

\begin{IEEEkeywords}
Computer Vision; Embedded devices; Location Detection; Pose estimation
\end{IEEEkeywords}

\section{Introduction} 

The unprecedented expansion of Internet of Things (IoT) devices and their advancing capabilities offer a perspective to the trend in computing for years to come, with more of the computations previously reserved for server-side now migrating to the edge of the Internet on resource-constrained devices. While this is now possible due to technology advancement, other factors also contribute to this accelerating trend, such as shifting perception at social and political levels, with users becoming more aware and concerned about their data privacy~\cite{jennifer2016user}. From policy makers, new legislation introduces heavy sanctions on companies for mishandling of users data~\cite{gdpr2018}, which pressures them to move more data processing to user proximity.

\begin{figure}[t]
\centering
\includegraphics[width=\linewidth]{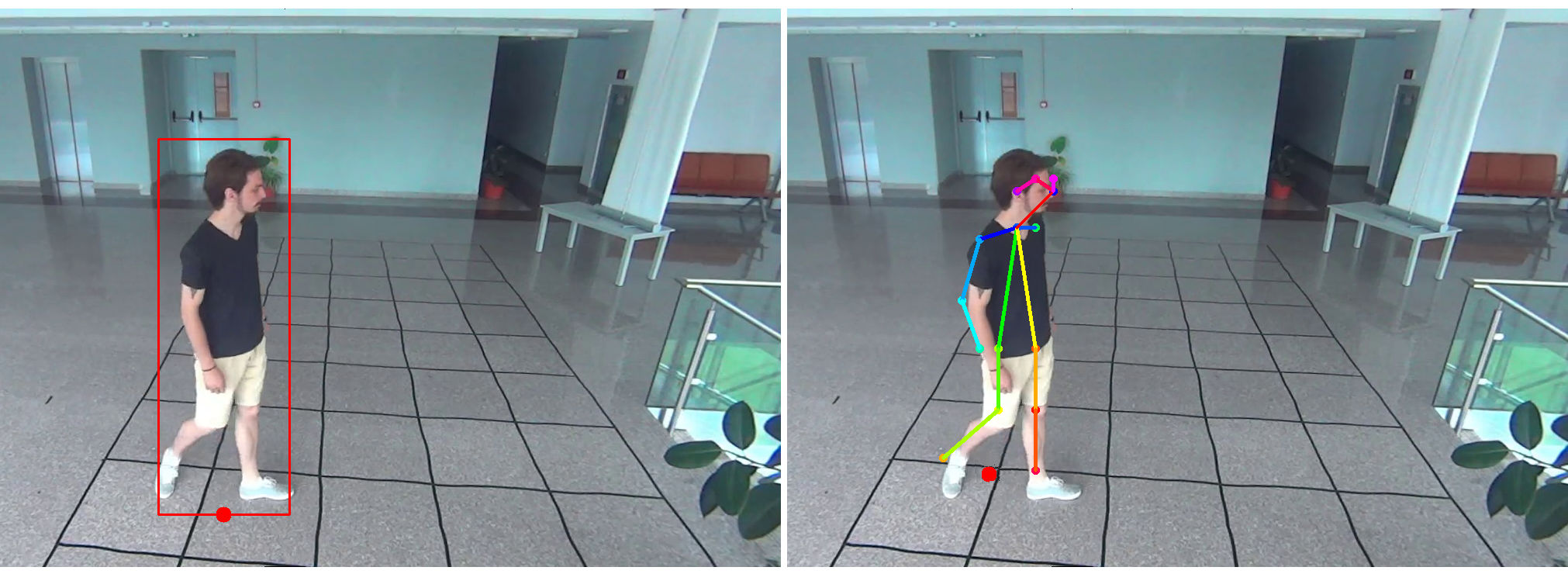}
\caption{Position estimation in the 2D-space in front of a camera indicated by the lower red dot using a bounding box approach (left) and our pose estimation approach (right). Our algorithm (CamLoc) achieves better detection accuracy with substantially lower compute resources.}
\vspace{-0.5cm}
\label{fig:detect_pose}
\end{figure}

Intelligent devices rely on sensors to understand user environment and context, to perform assistive actions and for user-device interactions. One of the richest information sensing modality is vision. A wide range of applications across different fields rely on vision for environment perception (surveillance, robotics and building automation and control). Knowing the exact location of a person in the environment in front of a camera is useful to many of these applications providing location based services. However, current methods used in computer vision require heavy computations, which are typically performed on servers with abundant resources. Migrating this detection task locally to a surveillance camera or other adjacent devices with limited embedded compute resources is not easy. Typically, algorithms designed to perform this detection on resource-constrained devices accept a severe downgrade of detection accuracy or inference time.

In this paper we develop a pose estimation algorithm building on key-points detection \cite{tome:3Dpose}, specifically designed to operate efficiently on embedded devices. To improve the location detection in different scenarios of occlusion, we extend the body frame determined from visible key body points to approximate the location of remaining key-points not visible in the image, enabled by appreciated body posture from visible points. We compare this with the performance of a popular detector YOLOv2 \cite{redmon:yolov2}, pruned to recognize only people, which determines a bounding box overlapping a person in image (as shown in Figure \ref{fig:detect_pose}). 

We collect a large dataset comprising images from cameras (over 2100 frames) annotated with the exact location of a pedestrian in the space in front of the camera. This dataset was designed to be particularly challenging for location detection, involving occlusions by having objects between pedestrian and camera to mask portions of the human body, in 42 different scenarios. As extension to just one camera, we consider situations where the algorithm can be improved by having access to a second camera facing the same scene from a different angle. We show that multiple cameras help to improve detection accuracy, assuming communication between cameras over the local network. This dataset will be made publicly available for other researchers to develop new algorithms for this challenging problem.

We show that CamLoc based on key body points detection performs better in comparison with a simple baseline relying on person detector with the popular YOLOv2 system for both inference time and detection quality.

This paper makes the following contributions:

\begin{itemize}
    \item We design a visual system for pedestrian location estimation in the 2D-space in front of a camera, building on key body points and based on posture estimation from the available body points extending pedestrian skeleton to compensate for occlusions and body outside of frame. This can work both in single-camera and multi-camera conditions.
    \item This approach is compared with a baseline based on bounding box generated by an object detection algorithm (YOLOv2), limited to just person detection. In order to cope with occusions, this is approach is adapted to extend bounding box such that it maintains the person ratio between height and width.
    \item We collect a substantial dataset annotated with pedestrian position in front of a camera to demonstrate the feasibility of our proposed technique. This includes both single camera and multi camera scenarios. We make this available for other researchers to propose new location estimation algorithms for smart-cameras, here: \href{https://bit.ly/2LzI8JE}{https://bit.ly/2LzI8JE}.
    \item Feasibility of running the two location estimation approaches on embedded devices is evaluated on Nvidia Jetson TX2 and Odroid XU4, which are two popular representatives of embedded devices for the IoT space. The deep neural networks in composition of these algorithms are evaluated on their resource requirements (inference time and memory footprint).
    
\end{itemize}

The structure of this paper is as follows. Next chapter introduces a brief motivation for the necessity of estimating pedestrian location in 2D-space. Section \ref{sec:techniques} presents the two detection approaches, followed by a presentation of our collected dataset. Section \ref{sec:evaluation}, presents the experiments to evaluate our proposed solution. Section \ref{sec:relwork} presents the Related Work. We finish with Future work and Conclusions (Section \ref{sec:conclusions}).

\section{Motivation}



There is a wide range of scenarios that require accurate localisation, some of which are highlighted by Mautz in \cite{mautz:indoor}: location based services in indoor environments, private homes e.g. Ambient Assistant Living (systems providing assistance to elderly people in their home for daily living activities), context detection and situational awareness, in museums (visitors tracking for surveillance and study of visitor behavior, location based user guiding and triggered context-aware information services), logistics and optimization (for the purpose of process optimization in complex systems, it is essential to have information about the location of assets and staff members), applications using augmented reality, and many other applications.

Most of the recent solutions in the area of indoor localisations that do not require specialized infrastructure use the sensors and WiFi cards in smartphones to determine user location. The solution proposed in this paper represents an alternative to these solutions, from the infrastructure perspective, which can be used in other scenarios as well, as it does not require carrying or having attached any device on the person being localized. As mentioned before, pedestrian localization is beneficial for tracking in surveillance and study of behaviour in museums, in shopping malls, in conferences, etc. These are attainable due to cameras already deployed for surveillance.

Two main aspects are to be considered for applications that make use of location data: low-latency interactions and data privacy. We are addressing both of these issues by running the detection on end devices, the cameras themselves with limited computation resources, rather than in the cloud.
This is possible due to the efficiency and low-resource utilization of the proposed deep neural network-based system (CamLoc).
Since surveillance is usually intended to be used as forensics rather than preventive, so far, we are not aware of other systems that perform detection on the cameras to offer real-time detection.

On resource-constrained end devices, such as IoT devices, it is desired to optimize resource consumption. This motivates our choices in designing a system that can operate on single frames at low frames per second. The techniques proposed perform detection on each frame in separation from other frames in order to cope with adaptable frame rate, dropping frames to save energy, depending on application requirements.

\section{Location Detection Techniques}\label{sec:techniques}

\begin{figure}[h]
\centering
\includegraphics[width=0.8\linewidth]{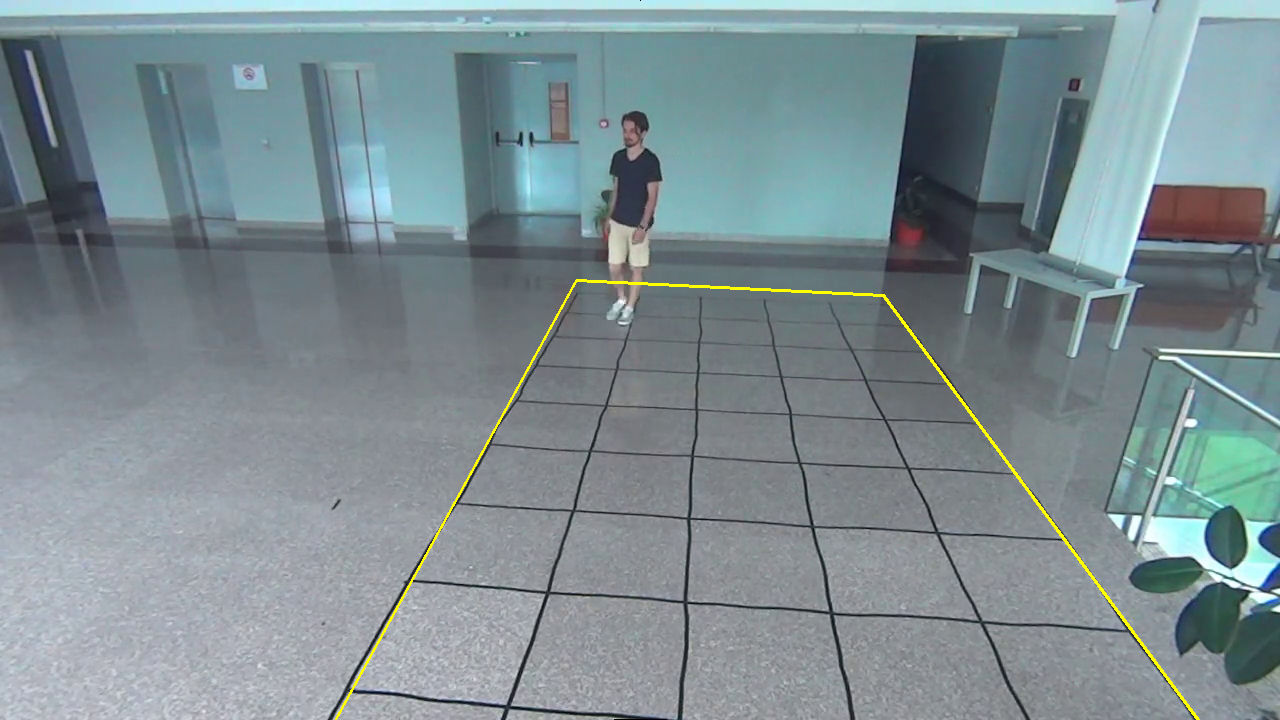}
\caption{Grid with homography points defined. Image has lens distortion corrected. Capture taken from S1\_Wide.}
\label{fig:homography}
\end{figure}

Given a camera with a view of the floor, localisation can be performed by using a homographic transformation from the position of the feet from the camera perspective to the floor plane (perspective-to-plane transformation). This method is based on the 2D direct linear transformation, developed by Abdel-Aziz and Karara \cite{aziz:indoor}. It implies that a set of 4 points must be defined a priori for a particular camera, as shown in Figure \ref{fig:homography}. The homographic transformation is based on the following formulae, given the camera perspective coordinates $x_c$ and $y_c$:

\begin{equation*}
X_{floor} = \frac{ax_c + by_c + c}{gx_c + hy_c + 1}
\end{equation*}

\begin{equation*}
Y_{floor} = \frac{dx_c + ey_c + f}{gx_c + hy_c + 1}
\end{equation*}

The parameters $a, b, c, d, e, f, g, h$ can be calculated by transforming the equations in matrix format, given the set of camera points $\{(x_i, y_i) \mid  i = \overline{0, 4}\}$, and a predefined set of map points $\{(X_i, Y_i) \mid i = \overline{0, 4}\} = \{(0, 0), (0, 1), (1, 0), (1, 1)\}$:

\begin{equation*}
    \begin{vmatrix} 
    x_1 & y_1 & 1 & 0 & 0 & 0 & -x_1X_1 & -y_1X_1\\
    x_2 & y_2 & 1 & 0 & 0 & 0 & -x_2X_2 & -y_2X_2\\
    x_3 & y_3 & 1 & 0 & 0 & 0 & -x_3X_3 & -y_3X_3\\
    x_4 & y_4 & 1 & 0 & 0 & 0 & -x_4X_4 & -y_4X_4\\
    0 & 0 & 0 & x_1 & y_1 & 1 & -x_1Y_1 & -y_1Y_1\\
    0 & 0 & 0 & x_2 & y_2 & 1 & -x_2Y_2 & -y_2Y_2\\
    0 & 0 & 0 & x_3 & y_3 & 1 & -x_3Y_3 & -y_3Y_3\\
    0 & 0 & 0 & x_4 & y_4 & 1 & -x_4Y_4 & -y_4Y_4
    \end{vmatrix}
    \cdot
    \begin{vmatrix}
    a \\
    b \\
    c \\
    d \\
    e \\
    f \\ 
    g \\
    h  
    \end{vmatrix}
    = 
    \begin{vmatrix}
    X_1 \\
    X_2 \\
    X_3 \\
    X_4 \\
    Y_1 \\
    Y_2 \\ 
    Y_3 \\
    Y_4  
    \end{vmatrix}
\end{equation*}

The problem then is how to estimate the position of the feet of a person in an occluded environment. For this, two methods have been analyzed:

\begin{itemize}
    \item Estimation using bounding box detection
    \item Estimation using pose information
\end{itemize}

\subsubsection{Baseline Person Detection}
Popular object detectors have good performance for the person detection task in terms of both accuracy and execution time (\cite{redmon:yolov2}, \cite{ren:fasterRCNN}). For our method, the technique described in \cite{redmon:yolov2} was used, having fast and accurate detections. It takes in a frame rescaled at the standard 224x224 resolution, and outputs coordinates of the bounding boxes and classes for each detected object in the frame. For our purposes, we are only interested in the \textit{person} class. The feet position can be estimated as the midpoint between the two bottom vertices of the bounding box. However, in occluded environments, this is problematic, as this method assumes the person is standing upright, and fully contained in the bounding box. Since the bounding box is much smaller than the height of the person and the image from the camera is at an angle, the feet position is erroneously estimated to be further from the camera than in reality. 

\subsubsection{Body extension using pose estimation}
This limitation motivated the use of pose information for inference. Since modern pose estimators (\cite{tome:3Dpose}, \cite{agarwal:pose1}) are able to detect subsets of body parts, this information is used to extend the person's body in the occluded area based on known body proportions \cite{doi:10.1080/03014460110085322}. This leads to an estimated feet position close to the actual one. Pose estimator neural architectures first detect person joints and through belief maps connect them to form body parts. Tome et al. \cite{tome:3Dpose} uses a multi stage convolutional neural network to output the pose information of a person. The network takes in a rescaled 224x224 frame and outputs the human skeleton. As such, the methodology for extending the human body is as follows:

    \begin{itemize}
        \item If the feet are found in the detections, take the point between them.
        \item Else, perform linear regression on the midpoint between complementary body parts (i.e. right/left shoulder, right/left hip) and extend onto the regressed line accordingly, considering the detected joints (e.g. extend the body starting from the lowest joint detected).
    \end{itemize}
    
Body extension is done on the regressed line from the detected joints, to account for natural body position (e.g. leaning against an object) and for eventual lens distortions. However, when insufficient joints are detected, regression cannot be performed and the frame is skipped. The percentages of skipped frames are shown in Table \ref{table:data_multi}.

\section{Human Position in Camera Frames Dataset}

\begin{table}
    \centering
    \begin{tabular}[h]{|c|c|c|}
    \hline
    Scene Name & \# scenarios & \# frames \\
    \hline
    S1\_Wide & 33 & 1929 \\
    S2\_Narrow & 9 & 267 \\
    \hline
    Total & 42 & 2196 \\
    \hline
\end{tabular}
    \vspace{0.3cm}
    \caption{Number of frames and scenarios in each scene.}
    \label{table:ds_frames_scenarios}
\end{table}

As such, the location detection is performed by first getting the video frames from cameras, running them through the deep learning model (either person detection or pose estimation), post-processing (extending the body, estimating the feet and, using camera configurations, averaging detections based on camera distance) and then computing global coordinates.

\begin{figure}[t]
\centering
\includegraphics[width=\linewidth]{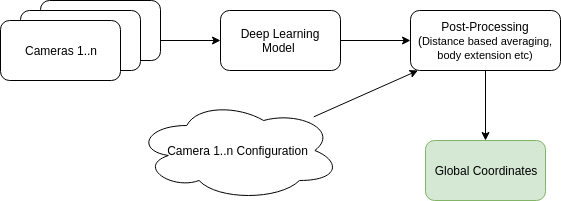}
\caption{Block diagram of the location detection methodology.}
\label{fig:method}
\end{figure}

To address this problem we start by collecting a wide dataset of video images annotated with exact location of person moving in a 2-D space in front of the camera. The collected dataset captures a single person in 2 different scenes from a total of 3 cameras. Each frame is annotated with the exact 2D position of the person in the scene. One of the scenes offers multiple points of view from 2 cameras simultaneously. Each scene is comprised of multiple localization scenarios, with varying levels of occlusion. A total of 42 scenarios are investigated across 2 scenes. A split down distribution of scenarios per each scene in presented in Table \ref{table:ds_frames_scenarios}. Each scene has an artificial grid drawn on the floor, that is used for validation. Global positioning is given relative to the origin of the grid.

\begin{figure}[t]
\centering
\includegraphics[width=\linewidth]{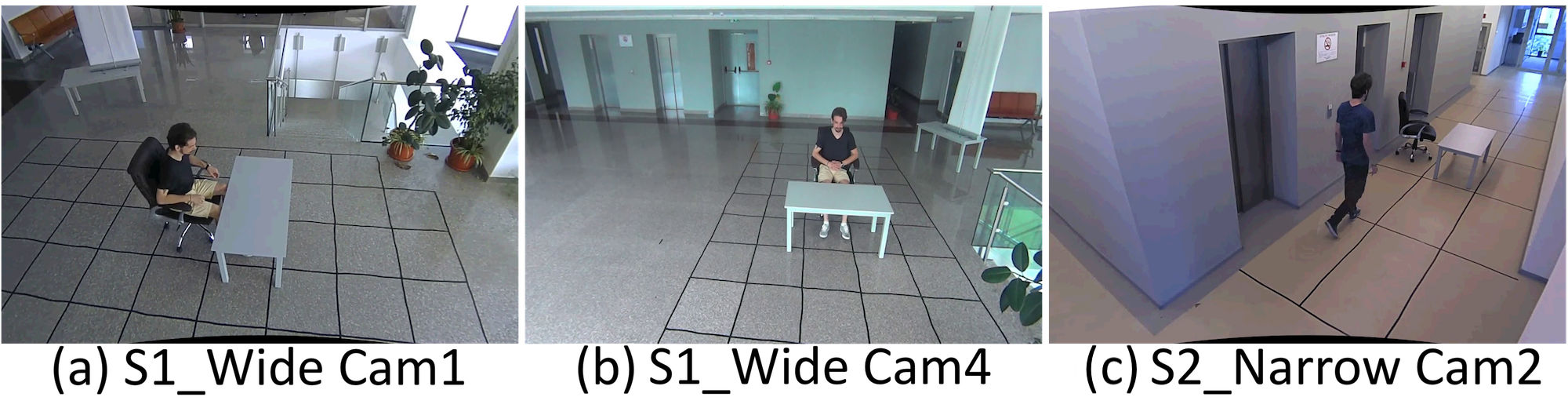}
\caption{Scenes and camera perspectives.}
\label{fig:ds_scenes}
\end{figure}

\subsection{Scenes Description}
The two scenes can be seen in Figure \ref{fig:ds_scenes}. Scenarios include obstacles at different distances and varying clothing.

\subsubsection{Scene 1} (Wide Space)
S1\_Wide represents a wide open-space such as a wide hallway, a lobby or a large room. Two camera perspectives are available at a perpendicular angle. These can be seen in Figure \ref{fig:ds_scenes} in images \textit{(a)} and \textit{(b)}. The two cameras are positioned at 2.8 meters and 1.8 meters, respectively, from the ground. The grid is a 540 cm x 300 cm rectangle, evenly divided into squares of 60 cm in length.

\subsubsection{Scene 2} (Narrow Space)
S2\_Narrow represents a narrow space, a typical hallway. The space reaches over 10 metres from the camera. The camera is at 2.5 meters from the ground, and the grid is a 225 cm x 1000 cm rectangle, divided into 75 cm x 90 cm rectangles. 

\subsection{Occlusions and Obstacles}

\begin{figure}[t]
\centering
\includegraphics[width=\linewidth]{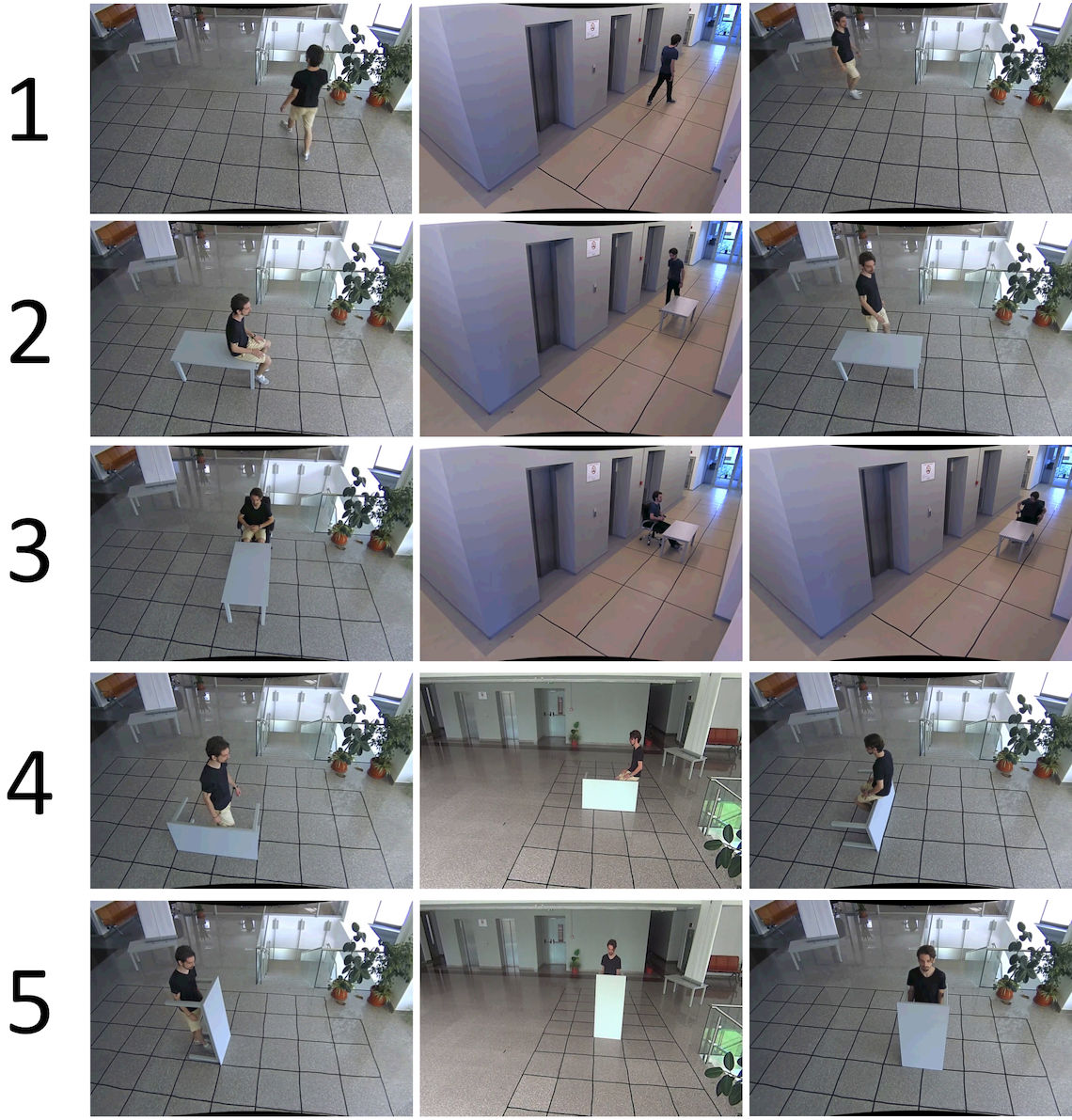}
\caption{Sample images from each scenario type (1 to 5). Not all scenes contain every type of scenario.}
\label{fig:ScenarioTypes2}
\end{figure}

The scenarios captured by the dataset can be grouped in 5 broad categories, described in Table \ref{table:scenario_types_description}. Situations with various levels of occlusions were considered, which could arise in real life scenarios. These include a person standing upright, sitting and with various body parts occluded by obstacles.
Sample images from each type of scenario are shown in Figure \ref{fig:ScenarioTypes2}. In some extreme cases, the body is almost completely covered (see scenario type 5. Table Standing), raising problems for vision-based positioning algorithms.

\begin{table}
    \centering
    \begin{tabular}[h]{|l|l|}
    \hline
    \textbf{Scenario Type} & \textbf{Description}\\ \hline
    1. Baseline & No occlusions present. This is the best case \\ & scenario \\ \hline
    2. Table & A simple table, used for testing localisation \\ &  when the person is sitting. \\ \hline
    3. Table and Chair &  A more complex variant of the previous type, \\ &  where feet are not always visible. \\ \hline
    4. Table Sideways & Used for occluding the lower part of the body.\\ \hline
    5. Table Standing & Occluding most of the person, except the upper \\ & part of the body.\\ \hline
\end{tabular}
    \vspace{0.3cm}
    \caption{Descriptions of scenario types across the scenes.}
    \label{table:scenario_types_description}
\end{table}

\subsection{Data Annotation}

\begin{figure}[h]
\centering
\includegraphics[width=0.8\linewidth]{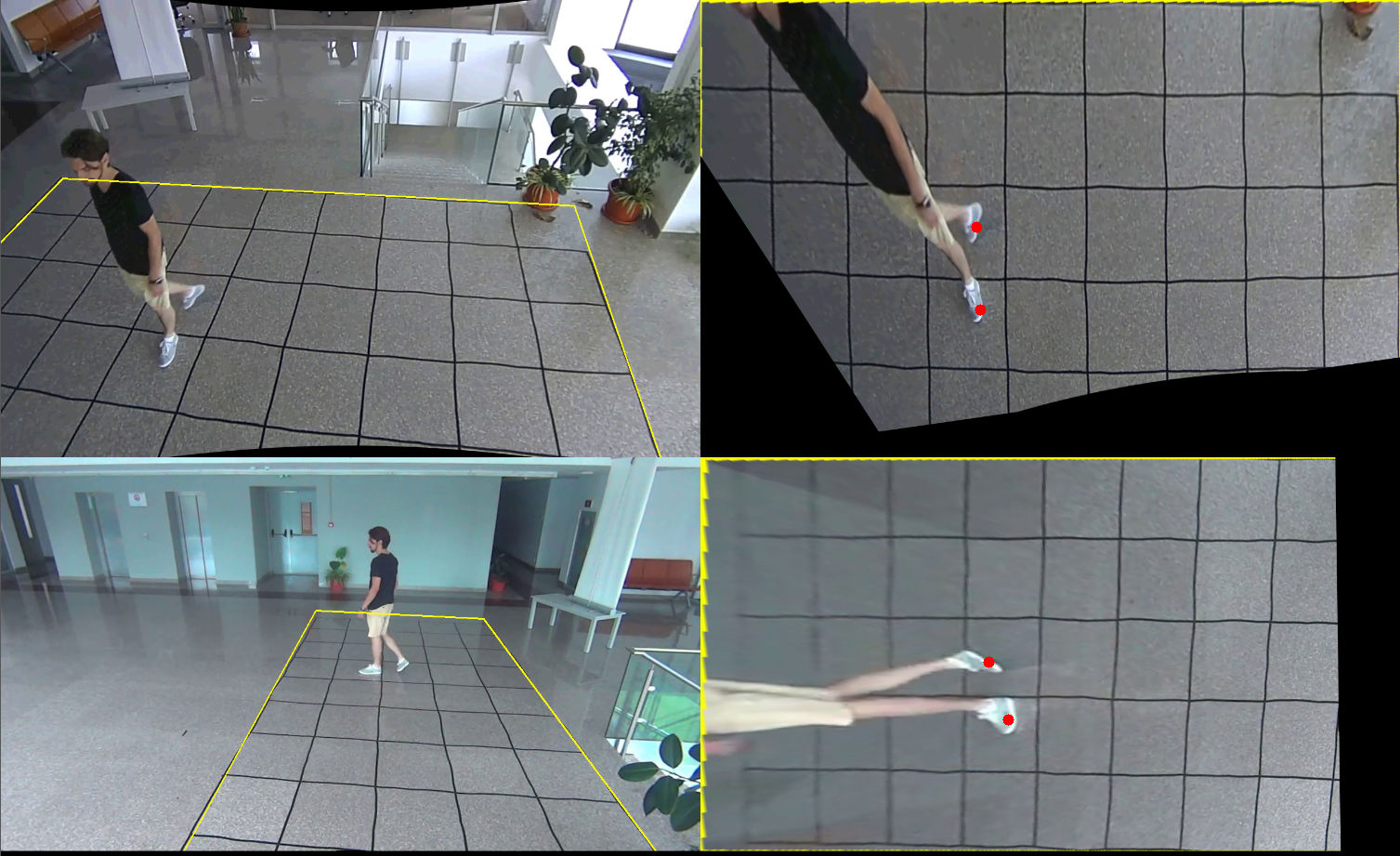}
\caption{Capture from the annotation tool, in the multi-camera scenario. Annotating the global location in this scenario requires annotating and combining both camera views. The red circles are the annotations on the transformed grid; the location of both feet are marked and then averaged to get the person location in one frame.}
\label{fig:anntool}
\end{figure}

For the S1\_Wide scene, the dataset offers the perspectives of two synchronized cameras. In this case, the ground truth annotation represents a combination of the annotations for the two camera perspectives: when the person is not visible on one of the cameras, the ground truth from the other camera represents the shared position. This approach is useful for situations when tracking the movement of people across video frames, including moving outside the coverage of one of the cameras. Otherwise, the position of the person is given by the midpoint between the annotations of the two perspectives. Figure \ref{fig:anntool} shows the annotation process.

\begin{figure}[h]
\centering
\includegraphics[width=0.8\linewidth]{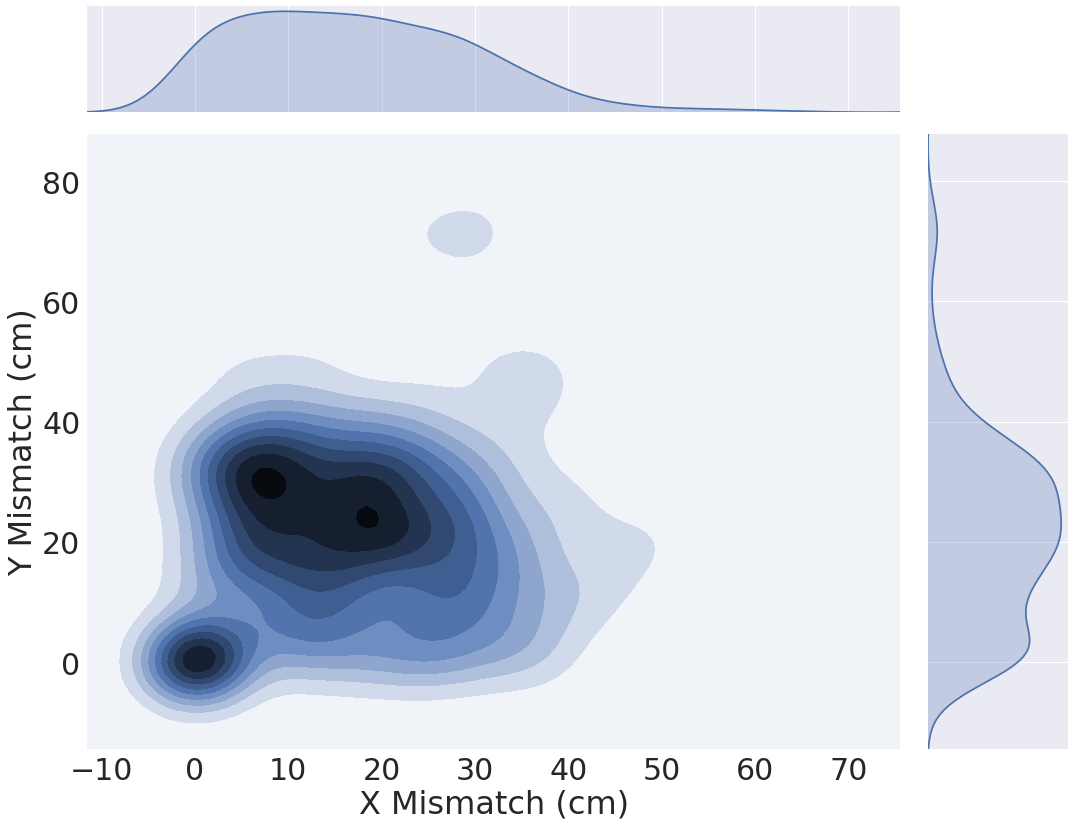}
\caption{Localisation mismatch in multi-camera perspective, with Y axis representing depth and X axis representing camera plain (95-percentile).}
\label{fig:jointerrs}
\end{figure}

Separately annotating two frames from different perspectives leads to different global coordinates. This is due to the differences in the set of points that define the homography, different frame rates and differences in synchronization. In the case of the S1\_Wide scene, which benefits from two camera perspectives, the localisation mismatch level is low, the average localisation mismatch for each axis being less than 20 cm. This is a good result considering that the distance between the person being tracked and the camera can reach over 6 m. The absolute differences in coordinates are presented in Figure \ref{fig:jointerrs}.

\subsection{Dataset processing}

The videos from the surveillance cameras were preprocessed beforehand to remove the barrel lens distortion. This was achieved with the use of a Linux's \textit{ffmpeg} command-line tool, \textit{defish0r}, which automatically corrects distortion, at the price of losing some information at the edges of the frame.

The frames were not scaled to a predefined set of dimensions. Instead, a configuration file is present for each scene, with the following information:

\begin{itemize}
    \item image height and width
    \item camera height and X,Y coordinates, with their respective units of measurement
    \item grid height and width, with units of measurement
    \item the set of points to define the homography transformation
\end{itemize}

Generally, the origin of the plane coordinate system is the lower left corner of the grid, as viewed by the camera. This is not the case for the multi-camera scenes, where the origin was chosen to be the same for both cameras. 

The dataset is offered as a set of frames from the gathered videos, with absolute X,Y coordinates annotations for each frame organized in \textit{.csv} files. 

The dataset was collected in a realistic environment from surveillance cameras \cite{Vivotek:dome, Vivotek:bullet} in an office building. It contains 2196 frames, and their distribution on each scene is shown in Table \ref{table:ds_frames_scenarios} (\textit{left}) .

\section{Evaluation}\label{sec:evaluation}

Evaluation is performed by analyzing the errors in localisation with respect to the ground truth annotations. Error is calculated as the euclidean distance between the global ground truth coordinates $p_{gt} = (x_{gt}, y_{gt})$ and the predicted coordinates $p_{pred} = (x_{pred}, y_{pred})$: 

$$d(p_{gt}, p_{pred}) = \sqrt{(x_{gt} - x_{pred})^2 + (y_{gt} - y_{pred})^2}$$

Considering that the predictors (object detectors / pose estimators) might not offer confident enough predictions for every frame, the percentage of missing predictions is also taken into account.

\begin{figure}[h]
\centering
\includegraphics[width=0.7\linewidth]{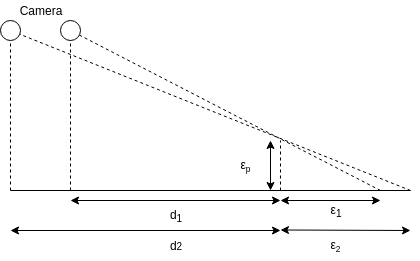}
\caption{Projection error with varying distance from camera}
\label{fig:triangles}
\end{figure}

Due to the way localisation is performed, by projecting a detection from the camera perspective onto the floor, the localisation error should have a positive correlation to the distance between the person and the camera. Using the properties of similar triangles, as shown in Figure \ref{fig:triangles}, maintaining the same camera height and varying the distance, leads to the following assertion:

\begin{equation*}
    \frac{\epsilon_2 - \epsilon_1}{\epsilon_1} = \frac{d_2 - d_1}{d_1}
\end{equation*}

This is to say that the relative increase of the localisation error is proportional to the relative increase of the person's distance to the camera. The error is also dependent on the predictor feet accuracy $\epsilon_p$ from the camera perspective. As such, with better predictor accuracy, the increase in localisation error is smaller. 

The fact that position errors increase with the distance from the camera shows that the methods presented in this paper are better suited for environments with good camera coverage.

\subsection{Detection in Single Images}

The person detection was performed using a pretrained YOLOv2 \cite{redmon:yolov2} object detector, with all classes discarded except the \textit{person} class. The YOLOv2 detector was trained on the VOC dataset \cite{pascal-voc-2012}, and is one of the most performant algorithms for object detection both in terms of accuracy and resource consumption. YOLO's custom \textit{Darknet} backbone architecture is one of the reasons for its speed, along with the fact that it belongs to the single-shot class of object detectors.

A gross estimate of the feet position is given by the centre of the lower edge of the bounding box, as shown in Figure \ref{fig:detect_pose}. In the case of a person standing without any body parts occluded, the bounding box method has good accuracy in estimating the feet position. It is a gross estimate because it does not work well in occluded environments, where the lower part of the body is missing (i.e. a person standing behind a table or a chair).

The optimization we use for the bounding box detection method consists of extending the bounding box to meet a particular aspect ratio - the aspect ratio of the bounding box when the person is standing. This is problematic since the mentioned aspect ratio depends on the camera height, the person's distance to the camera, and the person's orientation towards the camera, all of which cannot be known a priori.

Pose estimation was performed using the technique described by Tome et al. in \cite{tome:3Dpose}. The backbone architecture used is MobileNet \cite{howard:mobilenets}, which was chosen for its good trade-off between speed and accuracy. It makes use of depthwise separable convolutions for faster inference times. Lightweight neural architectures such as this one are becoming prevalent in the space of mobile applications. The network was trained on the COCO dataset \cite{lin:COCO}.

\begin{figure}[h]
\centering
\includegraphics[width=0.8\linewidth]{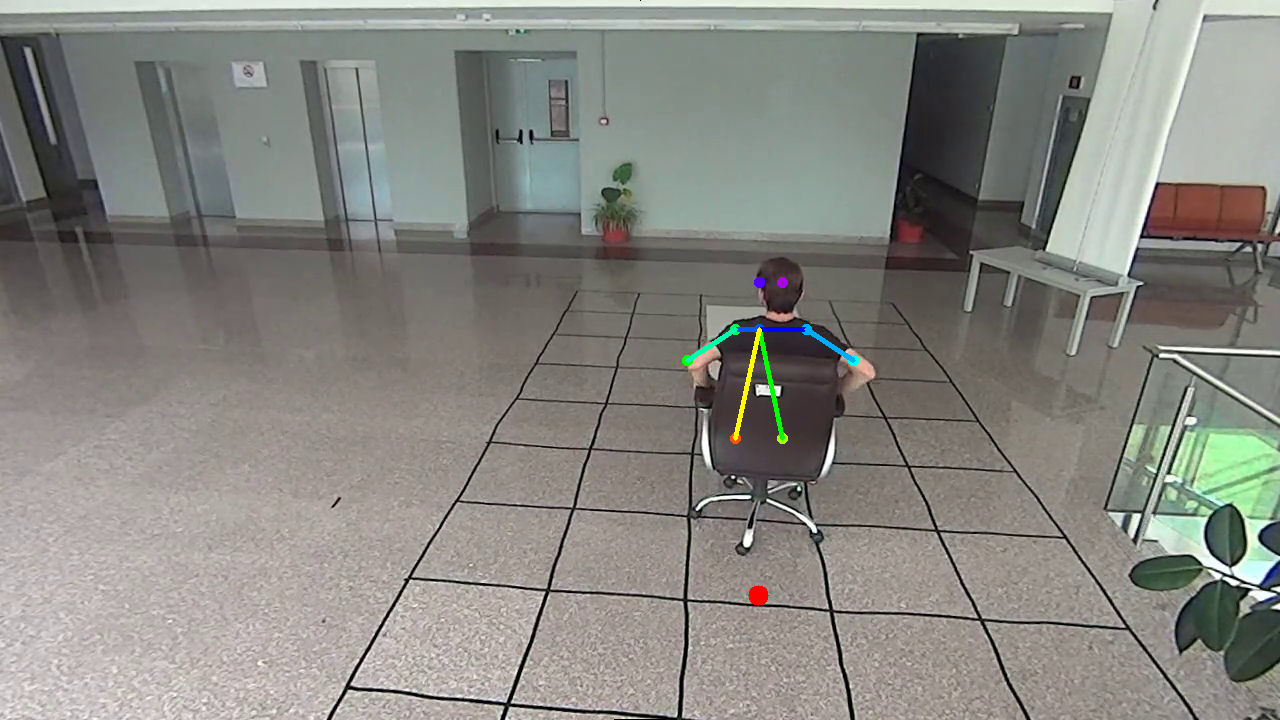}
\caption{Edge case when localisation with pose estimation fails to estimate the feet position.}
\label{fig:edge_case_pose}
\end{figure}

Pose information offers a more accurate estimation of the feet position, even in the cases when the feet detections are missing (see Figure \ref{fig:edge_case_pose}). The body position can be inferred from just a few body parts detected by using known body proportions. This is invariant to the camera position, since that information is contained in the estimated body proportion.

\begin{figure}[h]
\centering
\includegraphics[width=0.8\linewidth]{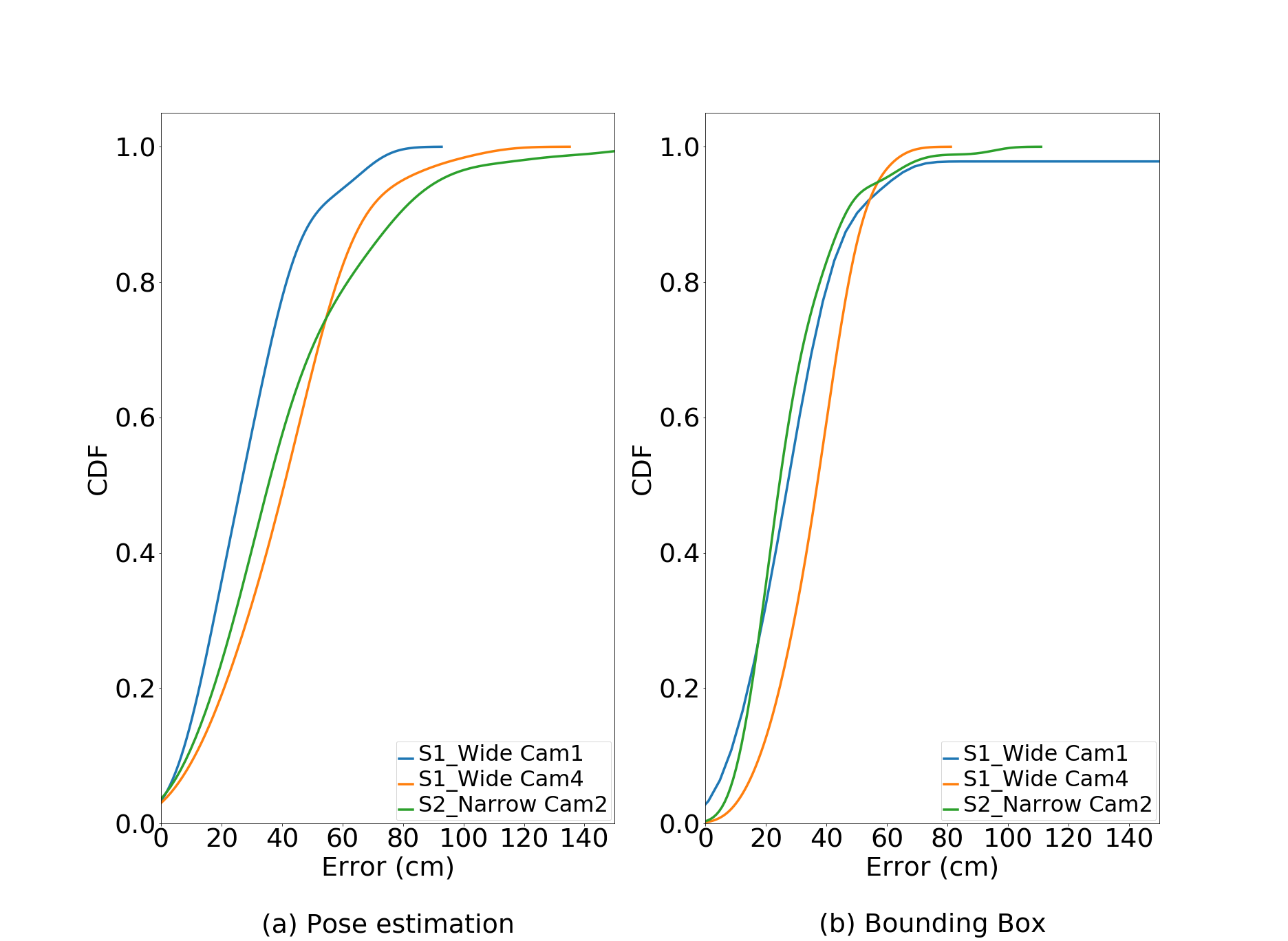}
\caption{Baseline scenario type error CDF.}
\label{fig:baseline}
\end{figure}

\begin{figure}[h]
\centering
\includegraphics[width=0.8\linewidth]{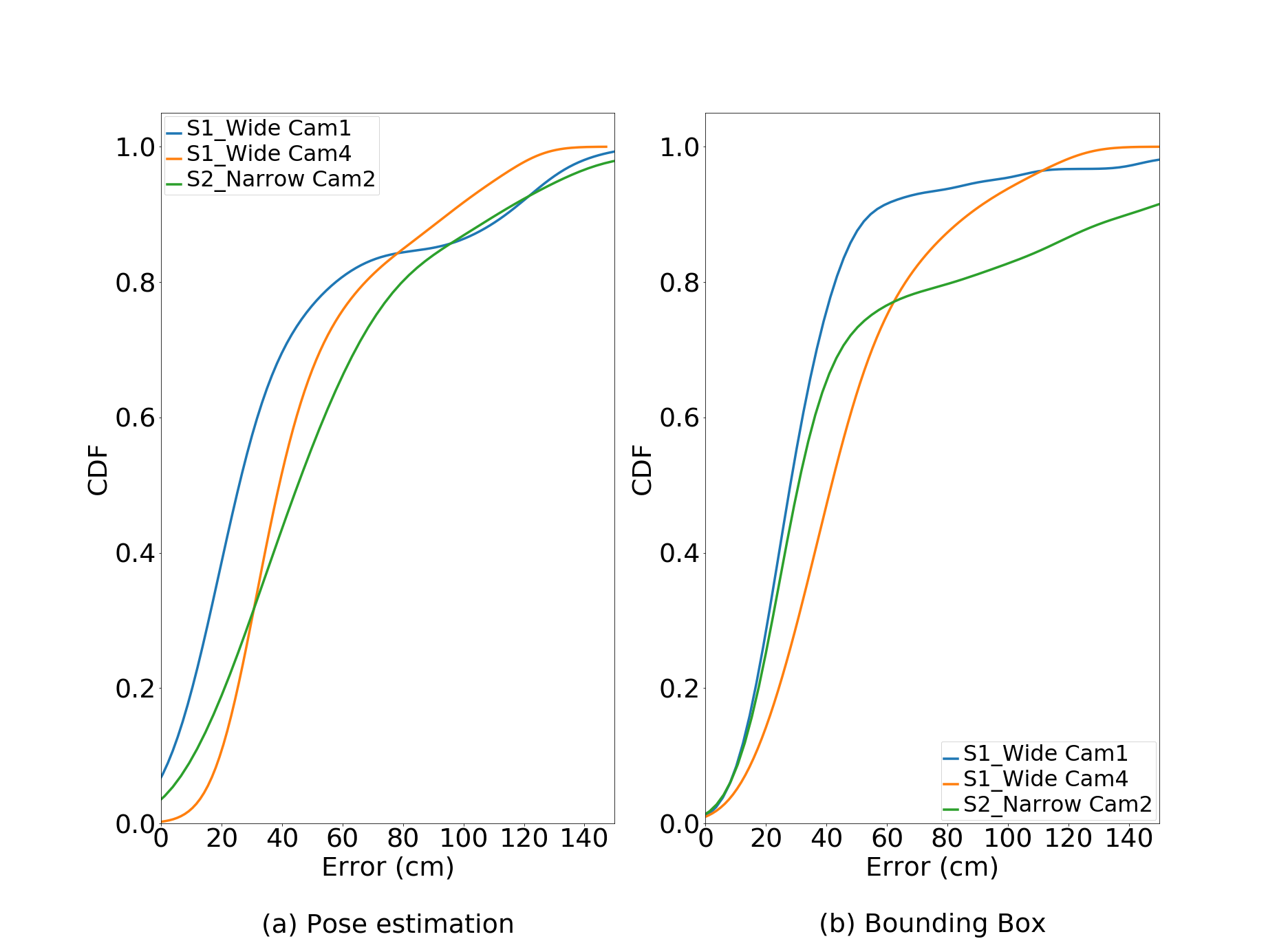}
\caption{Table scenario type error CDF.}
\label{fig:table}
\end{figure}

\begin{figure}[h]
\centering
\includegraphics[width=0.8\linewidth]{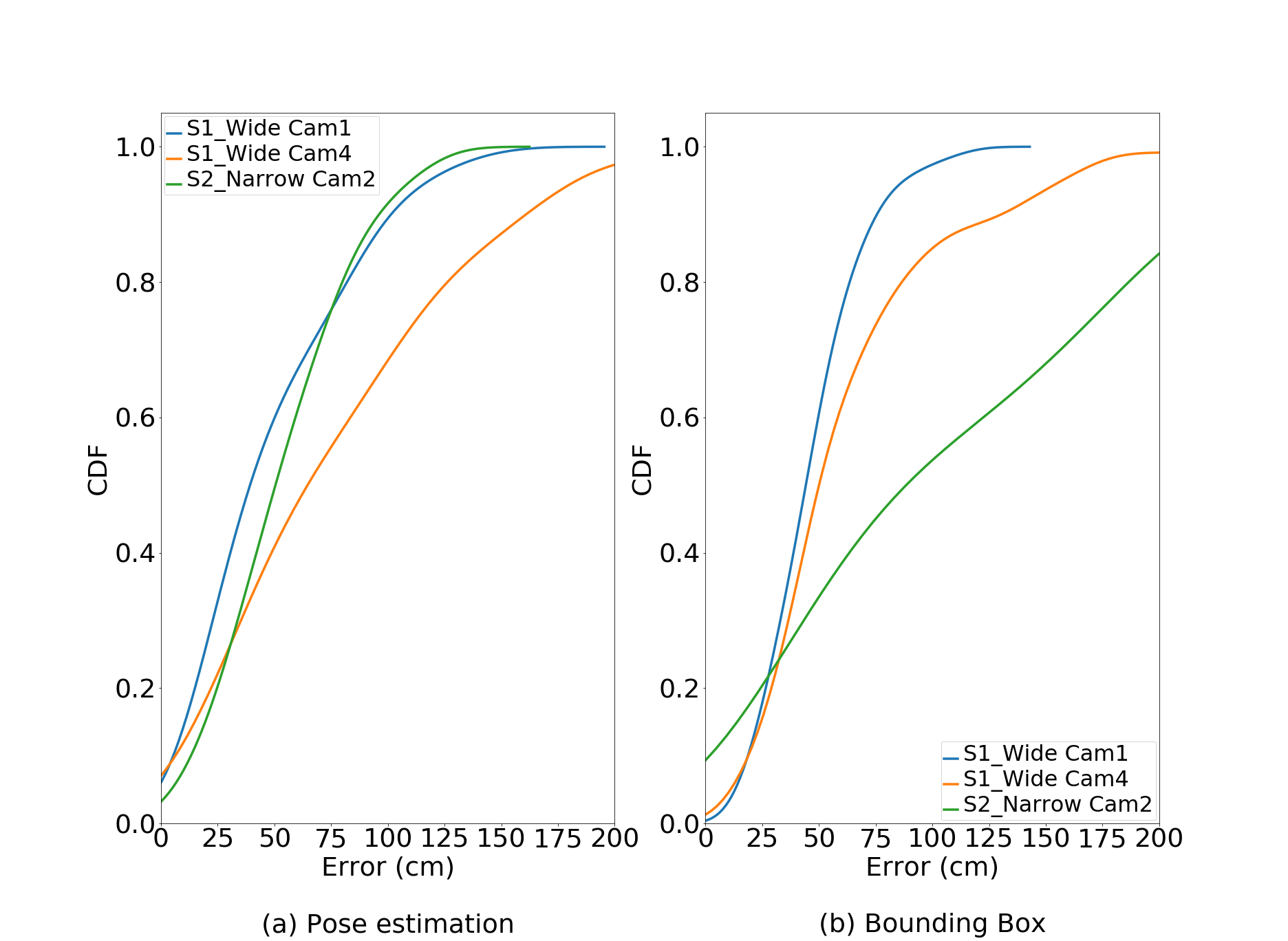}
\caption{Table and chair scenario type error CDF.}
\label{fig:table_chair}
\end{figure}

\begin{figure}[h]
\centering
\includegraphics[width=0.8\linewidth]{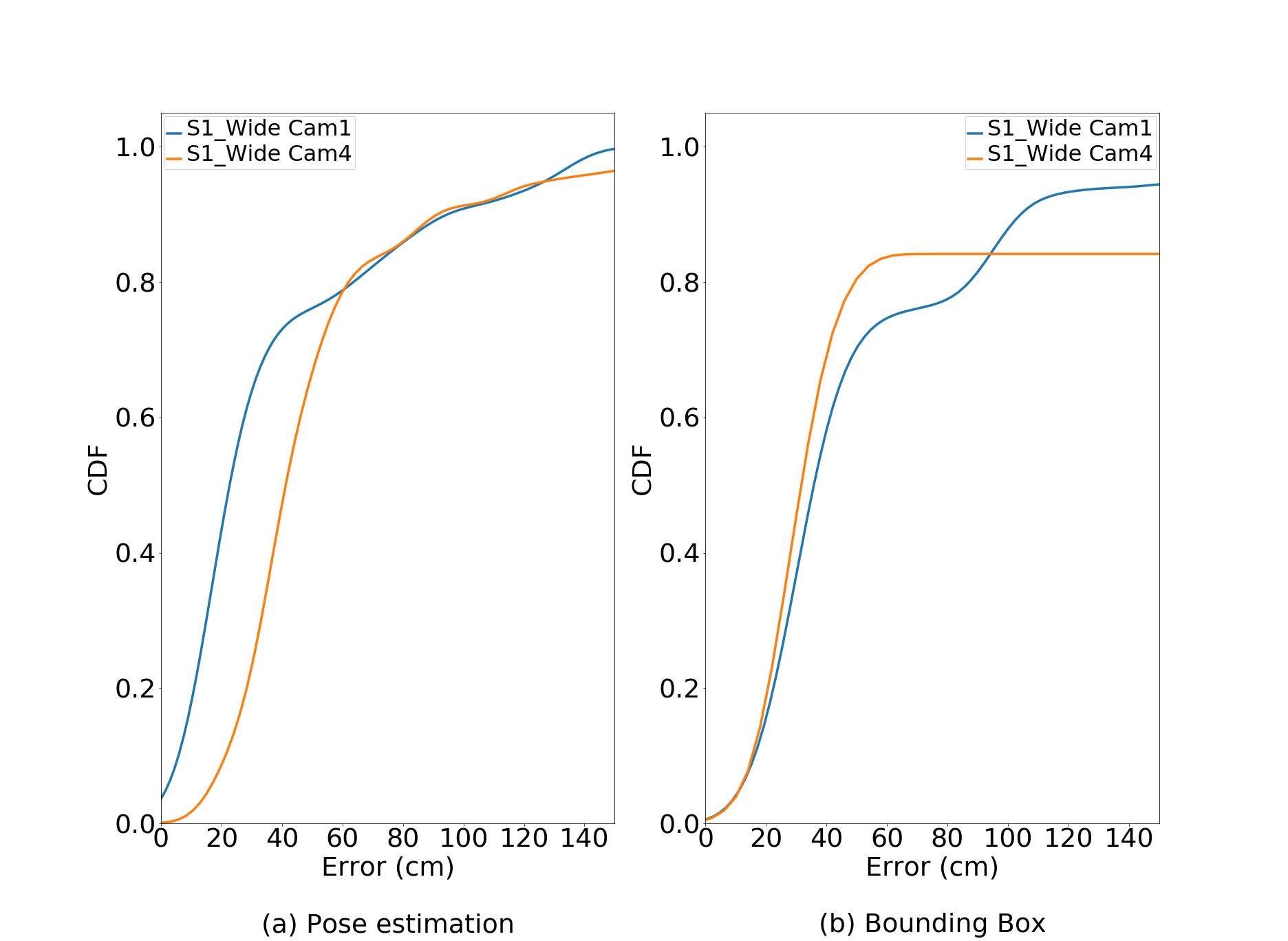}
\caption{Table standing scenario type error CDF.}
\label{fig:table_standing}
\end{figure}

\begin{figure}[h]
\centering
\includegraphics[width=0.8\linewidth]{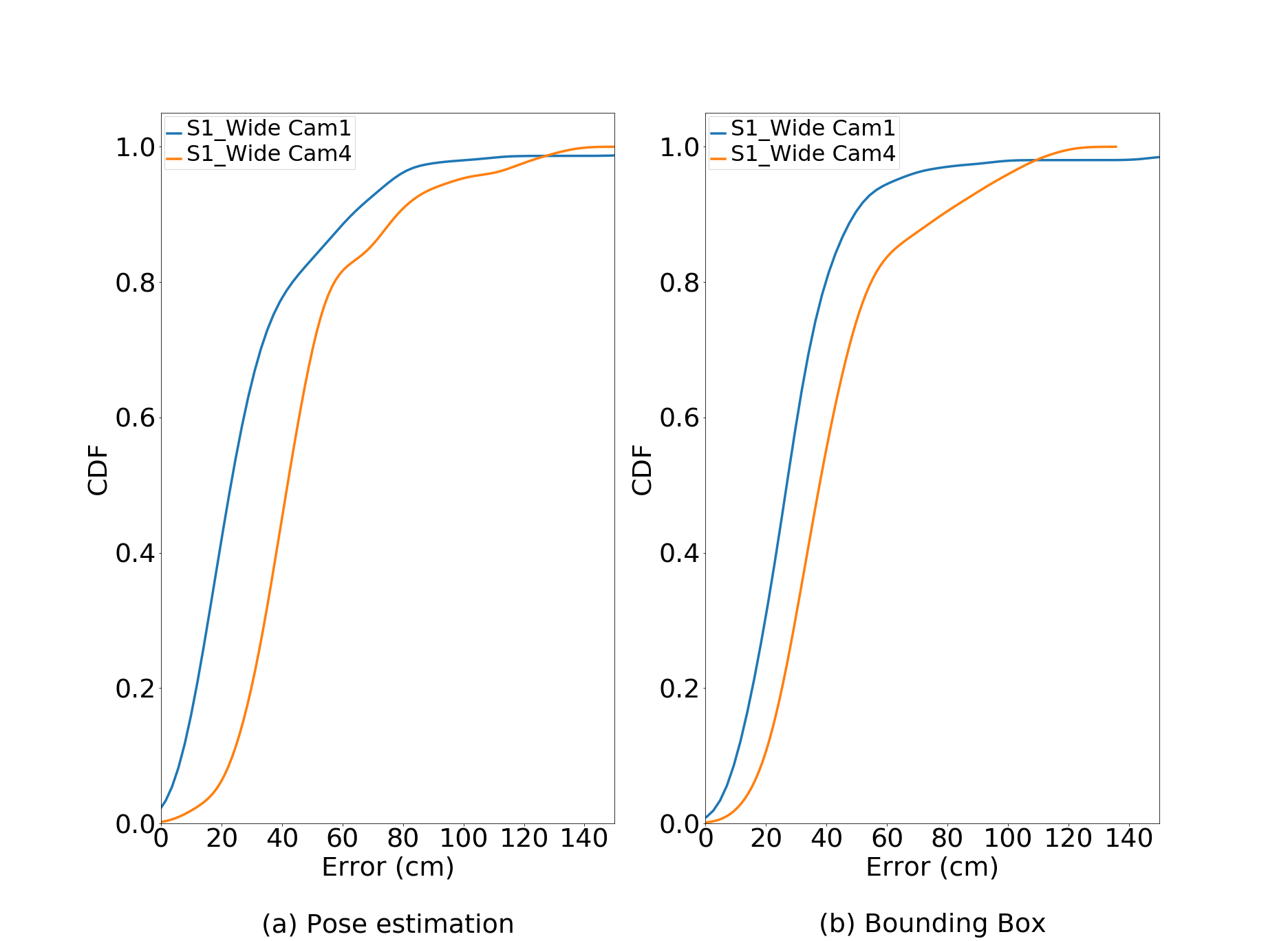}
\caption{Table sideways scenario type error CDF.}
\label{fig:table_sideways}
\end{figure}

\begin{table*}
    \centering
    \begin{tabular}[h]{|c|cc|c|}
    \hline
    \textbf{Scene} & \shortstack{\textbf{Pose estimation} \\ \textbf{mean error (cm)}} & \shortstack{\textbf{Bounding box} \\ \textbf{mean error (cm)}} & \shortstack{\textbf{Mean error} \\ \textbf{difference (\%)}} \\ 
    \hline
    S1\_Wide Cam1 & \textbf{36.26} & 41.99 & 13.6 \\
    S1\_Wide Cam4 & \textbf{53.58} & 60.99 & 12.1 \\
    S2\_Narrow & \textbf{45.27} & 48.37 & 6.4 \\
    \hline
\end{tabular}
    \vspace{0.3cm}
    \caption{Mean error value for both techniques (pose estimation and bounding box), using five cameras in three scenes.}
    \label{table:data}
\end{table*}

Table \ref{table:data} shows descriptive statistics of each scene, analyzed from a single camera perspective. In most situations, pose estimation has lower errors, and lower standard deviation compared to bounding box. Error cumulative distribution functions for each of the scenario types are presented in Figures \ref{fig:baseline}, \ref{fig:table}, \ref{fig:table_chair}, \ref{fig:table_standing}, \ref{fig:table_sideways}.

In these figures it can be observed that in the case of both methods, the position error is the lowest in the \textit{Baseline} scenario (CDF in Figure \ref{fig:baseline} and scenario in the first row of images of Figure \ref{fig:ScenarioTypes2}), where no occlusions occur. In this scenario, bounding box shows a slightly better performance than pose estimation. This is also the case in the \textit{Table} scenario (CDF in Figure \ref{fig:table} and scenario in the second row of images of Figure \ref{fig:ScenarioTypes2}) where the occlusions of the person are still minimal. However, when the occlusions are more significant (CDFs in Figures \ref{fig:table_chair}, \ref{fig:table_standing} and \ref{fig:table_sideways} and scenarios in the third, fourth and fifth rows of Figure \ref{fig:ScenarioTypes2}), pose estimation is outperforming bounding box.
In Figures \ref{fig:baseline} through \ref{fig:table_sideways}, it can also be noticed that in most cases, the lowest localisation errors are obtained by Cam1 in the S1\_Wide scene, most probably due to the position of the camera closer to the monitored scene.

Explanations for when pose estimation fares poorly consist of cases such as the one presented in Figure \ref{fig:edge_case_pose} when the body position is ambiguous, with no leg joints visible so the body position is interpreted as being upright. This case is representative for most of the bad predictions when almost all body parts are missing, and the body is interpreted as being upright, or in a position different from the actual one.

\begin{figure}[h]
\centering
\includegraphics[width=0.8\linewidth]{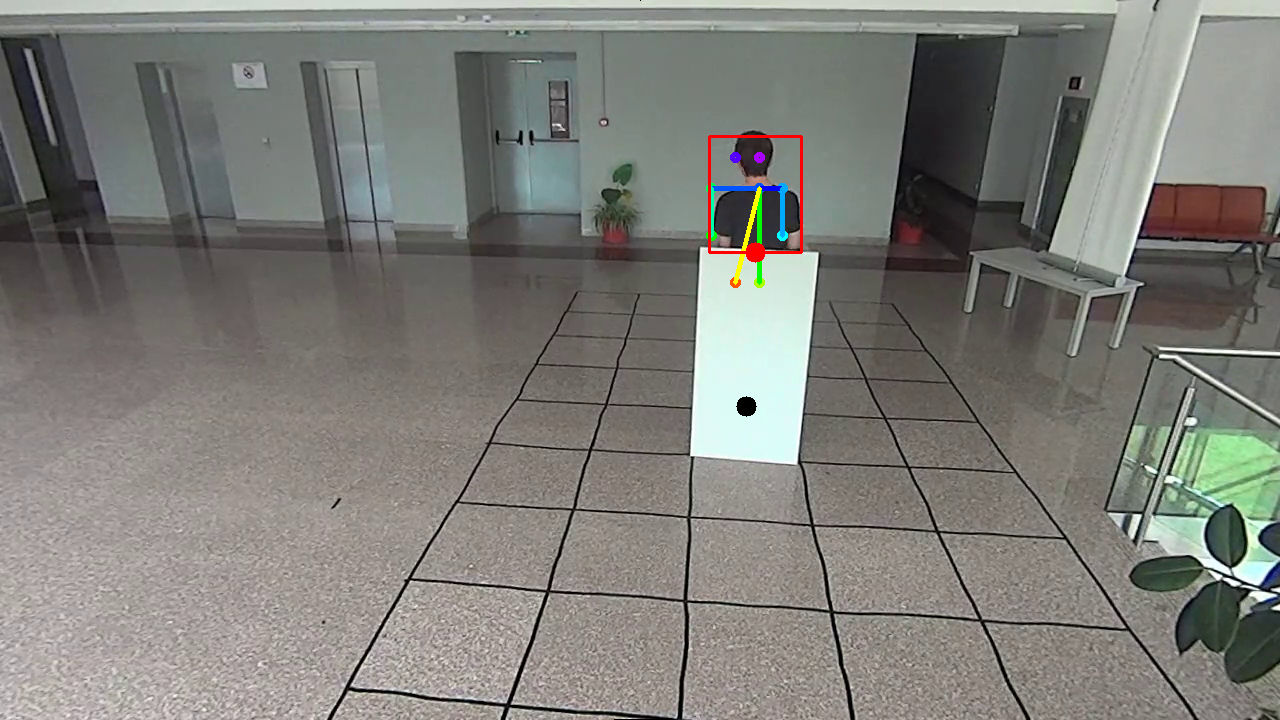}
\caption{Edge case when localisation with pose information works better than localisation with bounding box detections. The red circle represents the feet position offered by the bounding box detections. The black circle represents the feet position of the extended body to accustom for occlusions and missing body parts.}
\label{fig:edge_case_det}
\end{figure}

Bounding box detections also suffer from occlusion, but there is no real way to adjust the prediction: figure \ref{fig:edge_case_det} shows the case where bounding box fares poorly, while the body extension technique of the pose estimation method handles the occlusion well.

\subsection{Performance in Multi-Camera}

Considering the S1\_Wide scene, where positioning can be inferred from two different cameras at the same time, localisation could be improved by merging locations from both cameras using distance-weighted averaging:

\begin{equation*}
    P(p_1, p_2, d_1, d_2) = \left\{
        \begin{array}{ll}
              p_1 & p_2 \in \emptyset \\
              p_2 & p_1 \in \emptyset \\
              
              \ddfrac{d_2p_1 + d_1p_2}{d_1 + d_2} & \text{otherwise}
              
        \end{array} 
    \right.
\end{equation*}

\begin{figure}[h]
\centering
\includegraphics[width=0.8\linewidth]{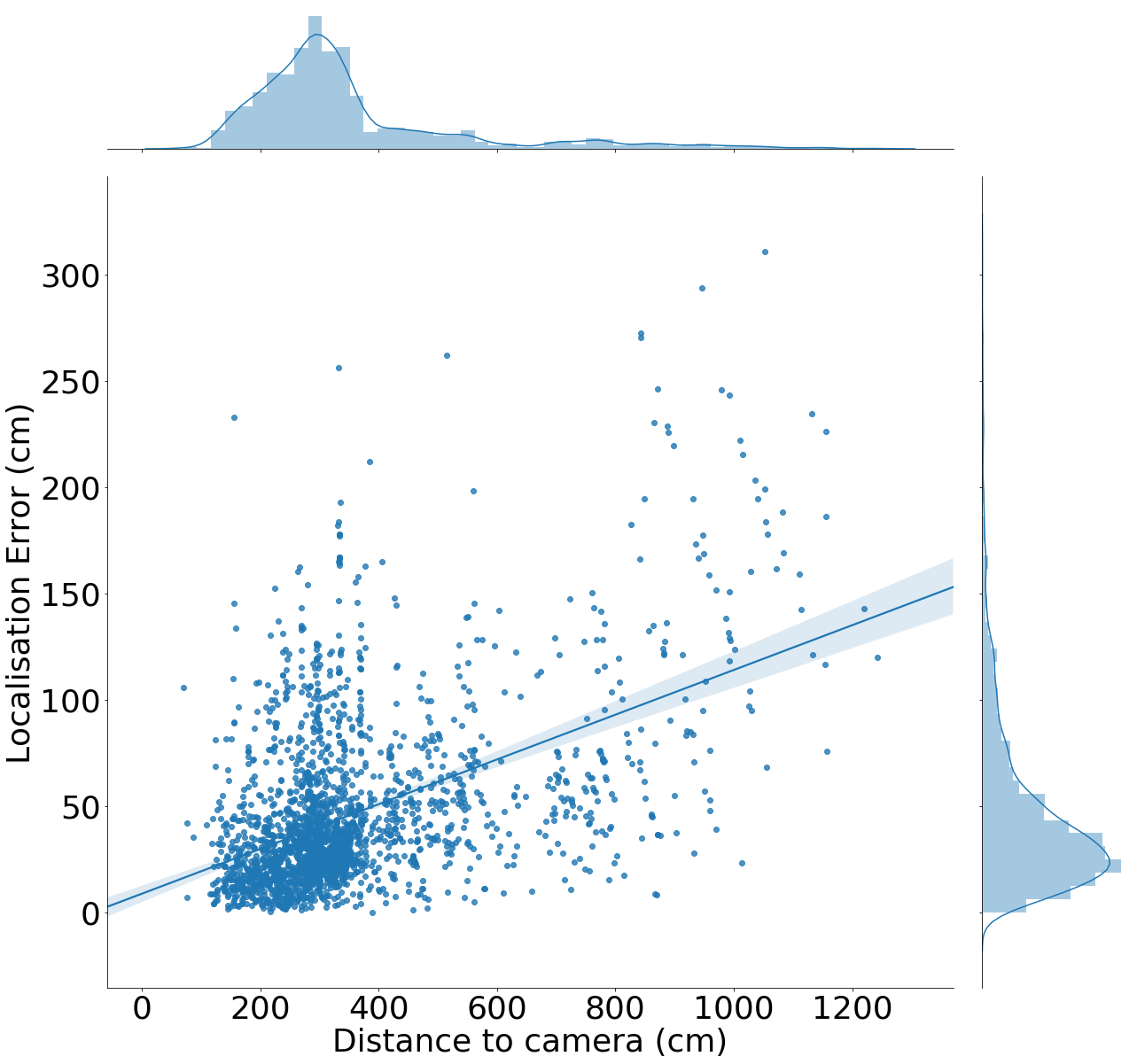}
\caption{Errors vs distance for pose estimation. Strong positive correlation.}
\label{fig:pose_dist}
\end{figure}

\begin{figure}[h]
\centering
\includegraphics[width=0.8\linewidth]{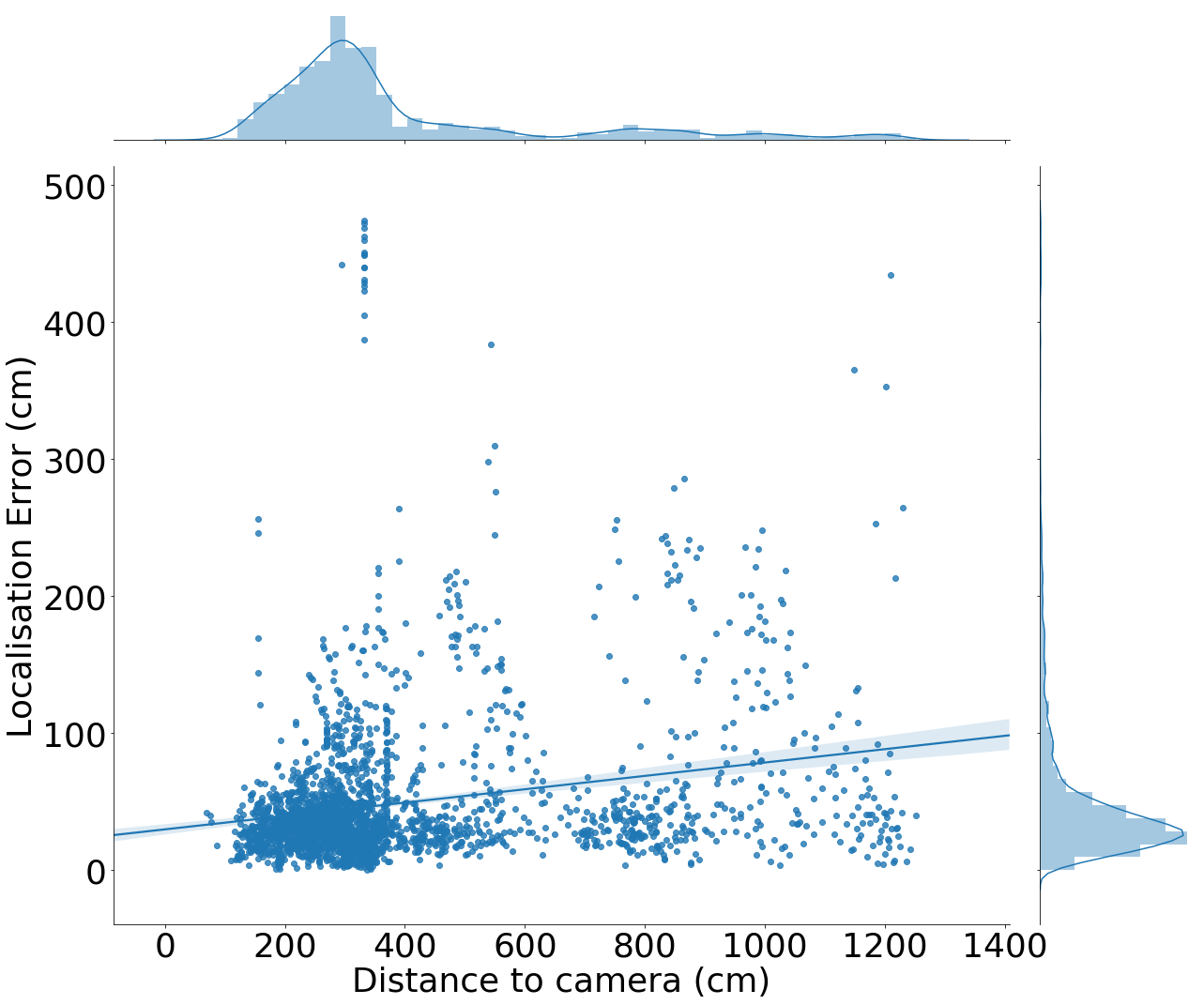}
\caption{Errors vs distance for bounding box. Correlation is not as strong as in the case of pose estimation.}
\label{fig:det_dist}
\end{figure}

The function $P$ takes into account the positions from both cameras and the distances to the camera. As such, when one of the cameras misses the prediction for a frame, the other camera supplies the position. If both cameras have inferred a position for the current frame, a weighted average of the two is computed using the inverse of their respective distances. This way, the position provided by the camera that is further away is penalized. This is motivated by the fact that localisation errors increase with the distance from camera, as shown in Figure \ref{fig:pose_dist} for pose estimation and Figure \ref{fig:det_dist} for bounding box.

\begin{figure}[h]
\centering
\includegraphics[width=0.9\linewidth]{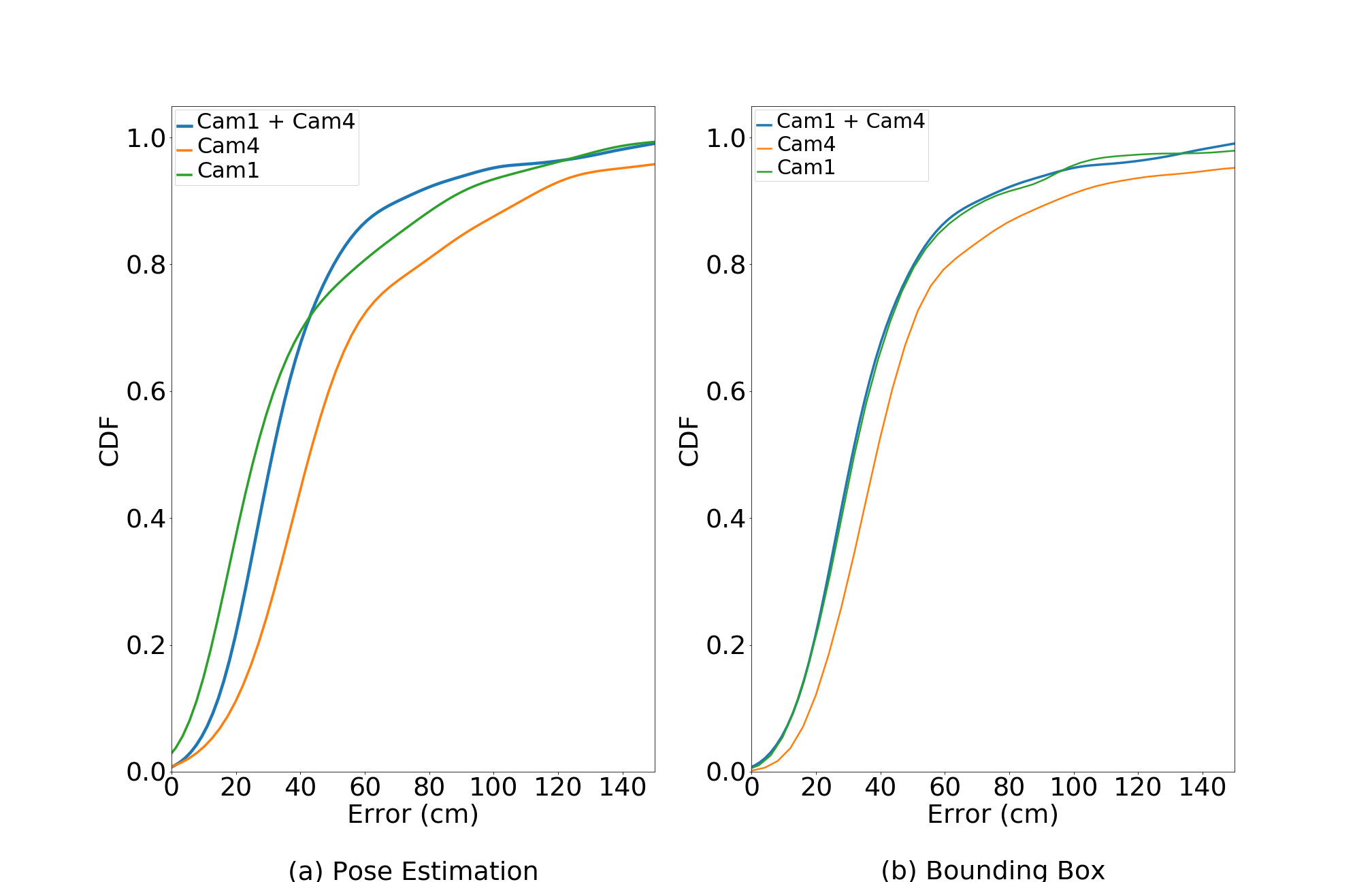}
\caption{CDF: multi-camera compared to individual cameras.}
\label{fig:c1c2multi}
\end{figure}

\begin{table*}
    \centering
    \begin{tabular}[h]{|c|cc|cc|}
    \hline
     & \shortstack{\textbf{Pose estimation} \\ \textbf{mean error (cm)}} & \shortstack{\textbf{Bounding box} \\ \textbf{mean error (cm)}} & \shortstack{\textbf{Pose estimation} \\ \textbf{missing predictions (\%)}} & \shortstack{\textbf{Bounding box} \\ \textbf{missing predictions (\%)}} \\
     \hline
    Cam1+Cam4 & 38.62 & \textbf{39.52} & \textbf{0.34\%} & \textbf{0.0\%} \\ 
    Cam1 & \textbf{36.26} & 41.99 &  9.18\% & 2.48\% \\ 
    Cam4 & 53.58 & 60.99 &  4.47\% & 0.33\% \\
    \hline

\end{tabular}
    \vspace{0.3cm}
    \caption{Performance results of multi-camera compared to individual cameras.}
    \label{table:data_multi}
\end{table*}

The perfomance of the multi-camera approach can be seen both in the CDF presented in Figure \ref{fig:c1c2multi} and in Table \ref{table:data_multi}. Since the procedure takes into account distances from both cameras, it can be noticed that errors smooth out. An important benefit of having multiple cameras consists is the significant improvement of the prediction ratios (less missed predictions) as shown in the right-hand side of Table \ref{table:data_multi}.

\subsection{Resources footprint}

\begin{figure}[!h]
\begin{center}

\subfloat[Jetson TX2]{\includegraphics[width=0.4\columnwidth]{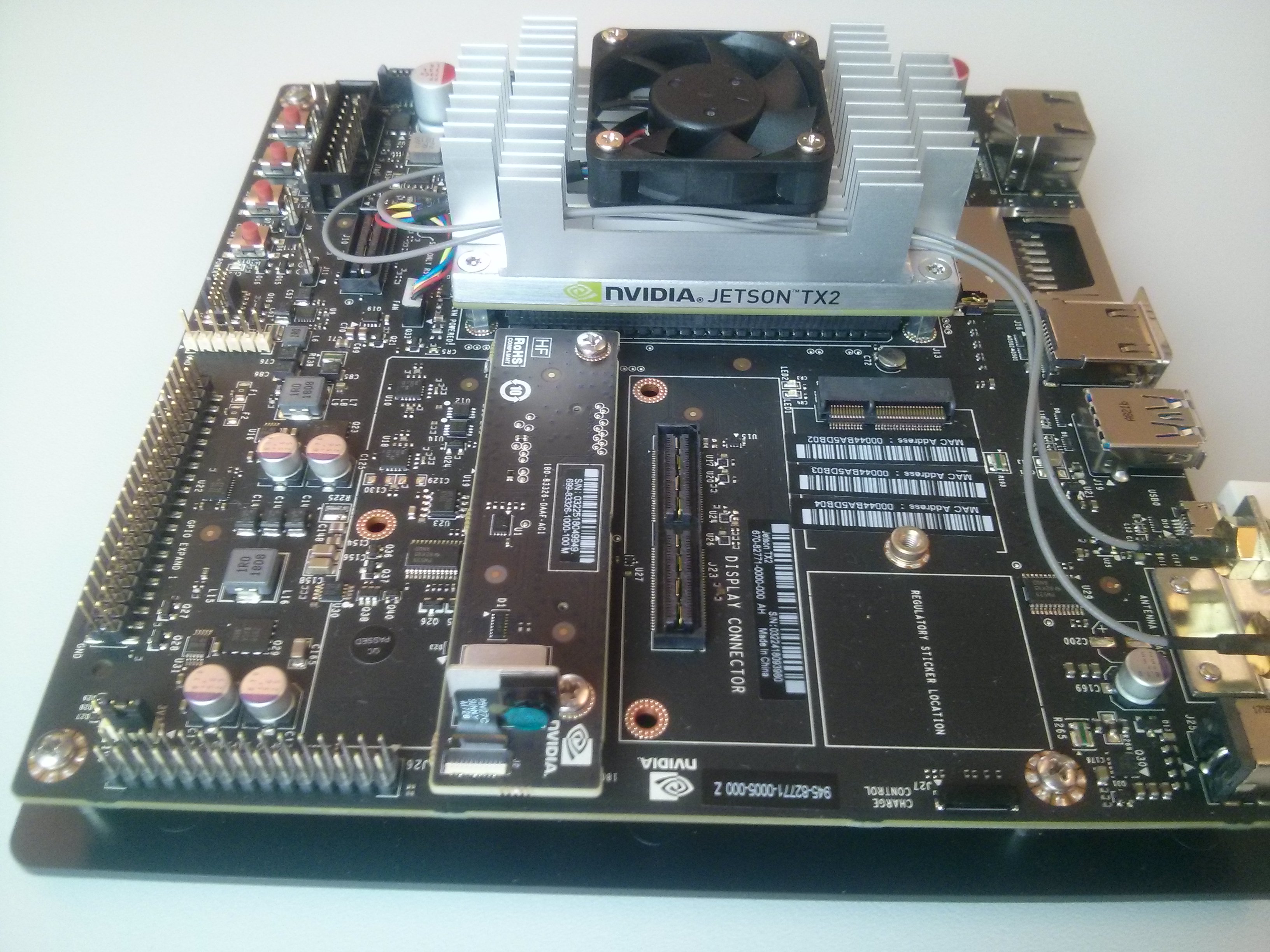}}
\qquad
\subfloat[Odroid XU4]{\includegraphics[width=0.4\columnwidth]{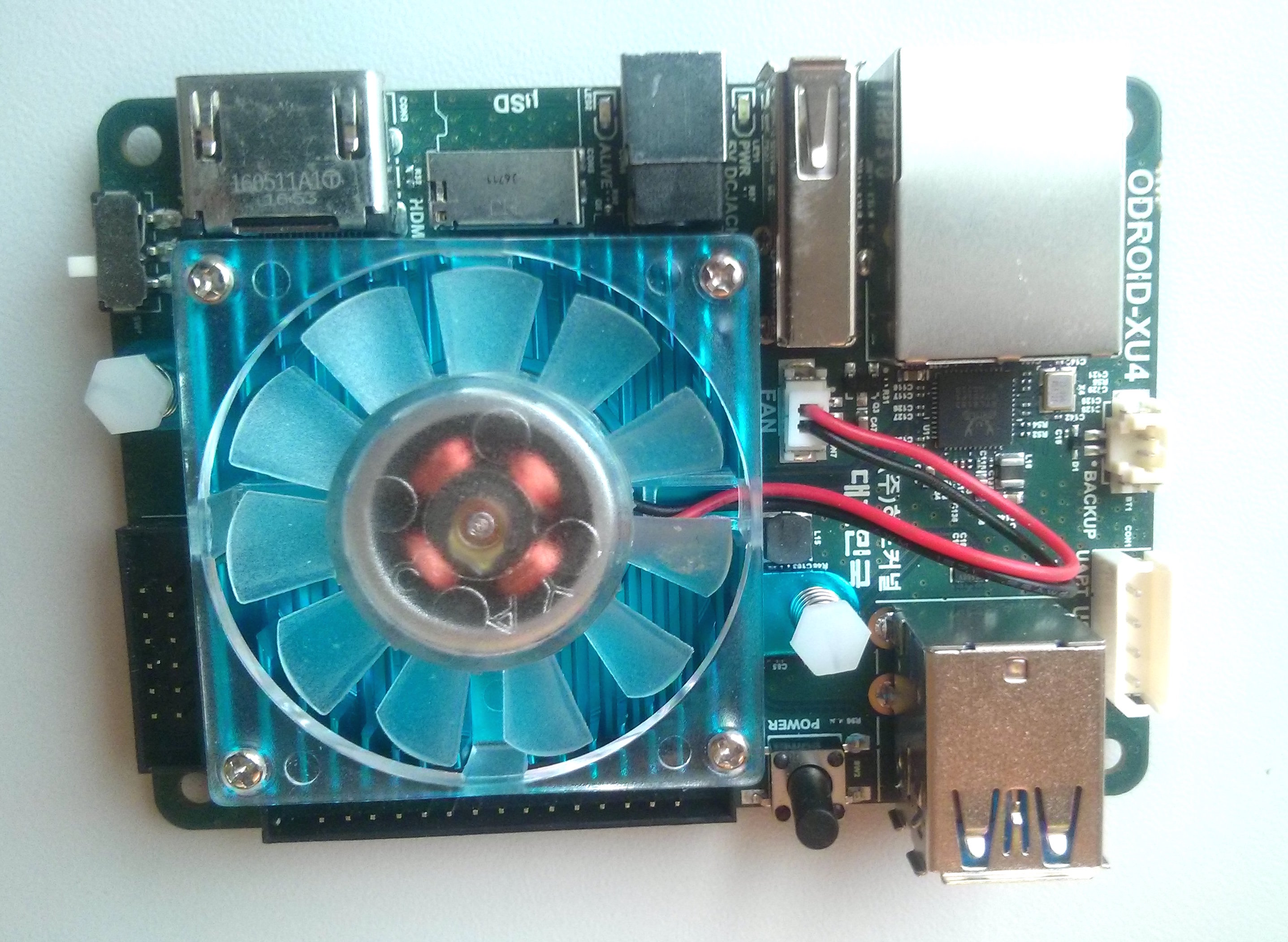}}
\caption{Experimented on resource-constrained devices.}
\label{fig:systems}
\end{center}
\end{figure}

We assess the performance of our proposed pose based localization system on two devices common to the embedded computing space, NVidia Jetson TX2 and Odroid XU4. The Jetson TX2 is a development platform with one integrated 256-core NVIDIA Pascal GPU, with 8GB of memory and a quad-core ARM Cortex-A57 CPU. The Odroid XU4 is an ARM big.LITTLE architecture, with four A15 and four A7 cores and 1.9GB of memory.

Table~\ref{table:on_devices} shows the inference time with batch size of one (one image at a time) and memory footprint, achievable at run-time on real hardware. The difference in memory footprint between Odroid and Jetson TX2 is due to the internal libraries used for Convolution computations by each of the two devices, Jetson relying on cuDNN, a highly optimized computation library for NVidia GPUs, maximizing speed in detriment to memory footprint, while the Odroid with ARM processor uses OpenBLAS, also a highly optimized matrix multiplications library, but agnostic to hardware profiles so balancing run-time memory and latency. Both of these exceed the baseline performance of the bounding box implementation on the Jetson CPU, in terms of both memory footprint and inference time. Due to memory constrains we were unable to run the baseline (bounding box) on the Odroid devices. Admittedly, the bounding box implementation based on YOLO is bulkier that necessary since the detection can handle multiple classes, here filtered just for the person class. A slimmer implementation of a person detector would yield different performance.

CamLoc on the Jetson TX2 achieves a frame rate of just above 6 frames per second. Even though is not enough to be classified as real-time performance for the human eye, it is still remarkably responsive for most applications that require location estimation for interactive services. 

\begin{table}
    \centering
    \begin{tabular}[ht]{|c|c|c|c|} \hline
     Device & Infer. time(sec) & Perform.(FPS) & Memory(MB) \\
     \hline
    Jetson TX2 (baseline) & 2.6 & 0.33 & 1520 \\ \hline
    Jetson TX2 (pose) & 0.16 & 6.25 & 620 \\ \hline
    Odroid (pose)& 0.45 & 2.22 & 210 \\ \hline
    \end{tabular}
    \vspace{0.3cm}
    \caption{Performance of baseline (bounding box) and pose estimation based localizations on NVidia Jetson TX2 and just pose estimation manageable on Odroid XU4.}
    \label{table:on_devices}
\end{table}

\section{Related Work}\label{sec:relwork}

\subsection{Object Detection}

One of the first methods to use deep convolutional networks in the context of object detection was R--CNN \cite{girshick:RCNN}. The approach was to extract region proposals using semantic segmentation and classify each of them using a SVM. As such, bounding boxes are generated with their respective classified class. Its main drawback was being slow, due each region being processed independently. Fast R--CNN \cite{girshick:fastRCNN} tries to reduce the execution time and memory usage by implementing region of interest pooling, more specifically Spatial Pyramid Pooling \cite{he:spatialPyramidPooling} to share computations. A final advancement to this method is Faster R--CNN \cite{ren:fasterRCNN}, which uses an "attention" model to propose regions through their Region Proposal Network.
However, even with the optimizations brought by Faster R--CNN, detection is not done in real time: Faster R--CNN registers 5 FPS using VGG net \cite{simonyan:veryDeep} with a mAP of 76.4. A faster but slightly less accurate approach is offered by YOLO \cite{redmon:yolo}, and its significantly more accurate successor, YOLOv2 \cite{redmon:yolov2}. The YOLOv2 network operates in real time, at 67 FPS using a modified GoogLeNet architecture with a mAP of 76.8. It frames detection as a regression problem, and predicts bounding boxes and class probabilities in a single evaluation. Since it removes the need of a detection pipeline (as in the spirit of Fast/Faster R--CNN), the system can be optimized as a whole.
Part of the larger class of single-stage detectors is RetinaNet \cite{lin:retinaNet} and it’s Focal loss function, which significantly increases accuracy. It was designed was to lower the loss for well classified cases, while emphasizing hard ones. Most of the time, two-stage detectors like Fast--RCNN tend to perform better accuracy--wise than single--stage detectors. This is due to single-stage detectors using a fixed grid of boxes, rather than generated box proposals. Still, RetinaNet has better performance on COCO dataset \cite{lin:COCO}.

\subsection{Pose Estimation}

Early approaches in estimating the pose of people \cite{agarwal:pose1, elgammal:pose2}, used direct mappings, HOG or SIFT, to build the pose from silhouettes. Nowadays deep learning approaches are ubiquitous, benefit from a large body of available datasets \cite{andriluka:poseDataset3, iqbal:poseDataset4, ionescu:poseDataset5}.
One of the most successful proposed approaches is DeepCut \cite{pishchulin:deepcut}, which initially detects people in the scene, and subsequently estimates their body pose. This approach uses a convolutional neural network for hypothesizing body parts and then performs non-maximum suppression on the part candidates. An improvement to this method was later compiled in the form of DeeperCut \cite{insafutdinov:deepercut}, which improved body parts detectors.
Another approach \cite{will:indoor5} uses a processing pipeline to first detect people in images and then estimate the pose. If the confidence of the detector is slim, pose estimation is skipped. Keypoints are predicted using heatmap regression with a fully convolutional ResNet. The system is trained only on COCO data, achieving state of the art results at the time. Tome et al., \cite{tome:3Dpose} propose an approach to detect 3D human pose. This method uses a 6-stage processing pipeline to "lift" 2D poses, using a combination of belief maps provided by convolutional 2D joint predictors  and projected pose belief maps.

\subsection{Vision-based Indoor Localisation}
Although different classifications for the existing indoor localisation solutions were offered throughout literature \cite{ez:indoor1, niki:indoor2, yang:indoor3, survey:indoor4, will:indoor5, rai:indoor6}, a simpler classifications divides them into solutions that require specialised infrastructure and solutions that make use of widely available infrastructure (such as wireless access points or surveillance cameras in buildings and inertial sensors in mobile devices).

Even though the majority of the existing indoor localisation solutions that make use of widely available infrastructure use smartphone sensors and WiFi to estimate the location, there has also been research into positioning systems by means of computer vision. These systems do not require users to carry special tags, enabling applications in circumstances where caring or wearing a tag is not viable (e.g. Ambient Assisted Living scenarios where the typical users are not well-versed when it come to technologically \cite{braun:ambiTrack}).

Mautz et al., have published a survey of optical indoor positioning system \cite{mautz:surveyCompVision}. The paper describes different systems and classifies them based on the reference used to determine the location of users in a scene such as images, projected patterns and coded markers. Tsai et al., propose in \cite{tsai:compVision2} a system that extracts from the video of surveillance cameras foreground objects using a background model. The system does not perform user identification, only positioning.
Several existing systems use RGB-D sensor for human positioning, such as the systems proposed by Munaro et. al., \cite{munaro:openPTrack} and Saputra et. al., \cite{saputra:compVision4} that offer a scalable multi-camera solutions for people tracking, Duque et al., \cite{duque:alaToate2017} who present a system for people localisation in complex indoor environments that works by combining WiFi positioning systems with depth maps, and the system proposed by Viola et al., \cite{viola:compVision6} that detects and identifies people, even if occluded by others, using an algorithm for creating free-viewpoint video of interacting people using hand-held Kinect cameras. Nakano et al., \cite{nakano:compVision7} present the potential applications for their proposed Kinect Positioning System, an indoor positioning system that uses Kinect without an infrared photophore.

Most of the positioning systems by means of computer vision use depth cameras, which cannot be considered part of the widely available infrastructure in large built environments. The localisation solution that we have proposed in this paper make use of typical surveillance cameras, that most large buildings are already equipped with.

\section{Future Work and Conclusions}\label{sec:conclusions}

Although the scenarios explored in this paper are complex in terms of the amount of human body occlusion and different postures, real-world applications many come with other forms of complexity. In future work we will explore scenarios with multiple people in the scene, which may impact the performance of body key-points detection and will require user identification.

The trend of performing more computations on IoT devices for local intelligences is likely to continue, with computer vision enabling a large class of applications that will migrate from cloud to the edge. Here we show that our system, CamLoc, based on human pose estimation can perform efficient location estimation, both in accuracy and in hardware resources footprint compared to YOLOv2 on single images from a fixed camera. Our annotated dataset also includes a multipe-camera perspective of the same scene, which contributes to improving detection when used in coordination. Our results show that such computer vision systems can operate efficiently on embedded devices, opening the opportunity for complex interactive applications in user environment assisted by smart-cameras to perform detections in user proximity.

\bibliographystyle{unsrt}
\bibliography{bib}

\end{document}